\pdfoutput=1

\documentclass[11pt]{article}

\usepackage[preprint]{acl}

\usepackage{times}
\usepackage{latexsym}

\usepackage[T1]{fontenc}

\usepackage[utf8]{inputenc}

\usepackage{microtype}

\usepackage{inconsolata}

\usepackage{graphicx}
\usepackage{booktabs}
\usepackage{adjustbox}
\usepackage{makecell}
\newcommand{\sys}{ARM}

\title{Can we Retrieve Everything All at Once? \\ARM: An \underline{A}lignment-Oriented LLM-based \underline{R}etrieval \underline{M}ethod}

\author{%
Peter Baile Chen$^{1}$ \quad Yi Zhang$^2$ \quad \textbf{Michael Cafarella}$^1$  \quad \textbf{Dan Roth}$^3$ \\
$^1$MIT \quad $^2$AWS AI \quad $^3$University of Pennsylvania\\
\small{Correspondence: \texttt{peterbc@mit.edu}}
}

\begin{document}
\maketitle
\begin{abstract}
Real-world open-domain questions can be complicated, particularly when answering them involves information from multiple information sources. 
LLMs have demonstrated impressive performance in decomposing complex tasks into simpler steps, and previous work has used it for better retrieval in support of complex questions.
However, LLM's decomposition of questions is unaware of what data is available and how data is organized, often leading to a sub-optimal retrieval performance.
Recent effort in agentic RAG proposes to perform retrieval in an iterative fashion, where a followup query is derived as an action based on previous rounds of retrieval. 
While this provides one way of interacting with the data collection,
agentic RAG's exploration of data is \emph{inefficient} because 
successive queries depend on previous results rather than being guided by the organization of available data in the collection.
To address this problem, we propose an LLM-based retrieval method --- \sys{}, that aims to better align the question with the organization of the data collection by exploring \emph{relationships among data objects} beyond matching the utterance of the query, thus leading to a retrieve-all-at-once solution for complex queries.
We evaluated \sys{} on two datasets, Bird and OTT-QA. On Bird, it outperforms standard RAG with query decomposition by up to 5.2 pt in execution accuracy and agentic RAG (ReAct) by up to 15.9 pt.
On OTT-QA, it achieves up to 5.5 pt and 19.3 pt higher F1 match scores compared to these approaches.

\end{abstract}

\section{Introduction}~\label{sec:intro}
\begin{figure*}
\centering
\includegraphics[width=\linewidth, trim=0 1.8cm 0 0, clip]{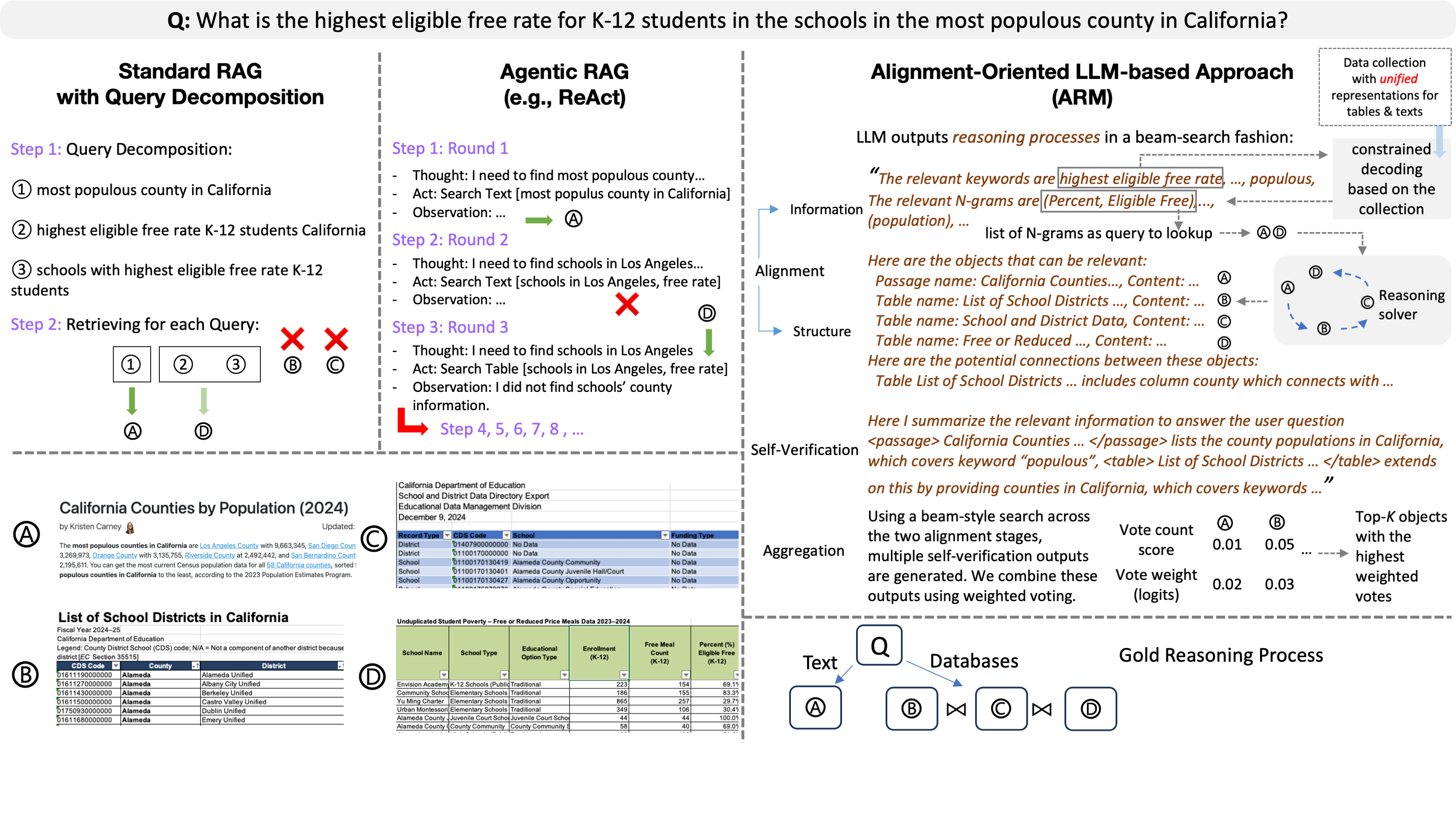}
\caption{A summary of our approach \sys{}, and a comparison with retrieval in standard RAG, which leverages LLMs for query decomposition, and agentic RAG, which employs LLM-based agents to iteratively generate queries.}
\label{fig:problem}
\end{figure*}

Answering real-world questions can be complicated, especially when required information is distributed across \emph{heterogeneous} information sources, such as text corpus, databases, and even image collections.
Consider an example question \textit{"What is the highest eligible free rate for K-12 students in the schools in the most populous county in California?"} and the data collection shown in Figure~\ref{fig:problem}. Answering this question requires one passage (A) from the text corpus and three joinable tables (B, C, D) from the database.
To find all of these passages and tables, it
requires exploring the available text and tables in the data collection, reasoning about their relationships (e.g., joinable tables, entity connection), and determining the best organization of these objects that can fully answer the question.

With recent advances in LLMs, they have been used to decompose complex questions to boost the performance of RAG systems~\cite{khot2022decomposed,zhou2022least,jhamtani2023natural}. However, all these decomposition methods are unaware of what data is available, and how they are organized. Therefore, these approaches are highly likely to overlook important data objects relevant to the question, especially when their information is not explicitly mentioned. As depicted in Figure \ref{fig:problem}, this method can miss all bridging tables.

A possible way to improve LLM's decomposition is to make it interact with the data collection, which has been explored in the context of agentic RAG~\cite{react,ircot,self-ask-2023,self-rag-2023,zhang2024agentic}. Typically, an LLM-based agent, which has demonstrated impressive performance in reasoning and acting~\cite{kojima2022large,wei2022chain,shinn2024reflexion,yao2024tree}, \textit{iteratively} reasons about actions (i.e., search queries). Search queries are then issued, and the retrieved passages will be fed back to the agent to reason about the next action (search query) until it determines that the question has been fully answered.
Although we have seen promising results when adapting such agent-based iterative solutions, they have shortcomings as follows:


First of all, although the iterative approach enables interaction with the data collection, each action is guided by the agent's decision about what is still missing to fully answer the question based on the information retrieved in previous rounds, rather than what data objects are available to answer the question.
Therefore, it can lead to an \emph{inefficient} exploration with unnecessary LLM calls, increasing inference time and cost.



Secondly, an iterative approach based on past experience can not enable a \emph{joint optimization}, where earlier and later steps are planned together.
As shown in ~\cite{react}, an agentic solution faces a critical issue known as \emph{reasoning derailment}, a phenomenon in which
agents struggle to recover from an initial erroneous action, resulting in a further round of incorrect reasoning.

In this work, we argue that enabling an efficient and comprehensive retrieval for complex questions requires aligning the question and its decomposition to existing data objects and their organization.
This means that during the retrieval process, in addition to performing semantic matching between the question and data objects, we need to simultaneously reason about how candidate data objects can be connected and whether they involve bridging entities or tables. 
Therefore, we propose an LLM-based \textbf{a}lignment-oriented \textbf{r}etrieval \textbf{m}ethod, \sys{}, for complex open-domain questions. Through reasoning about a better alignment to the data collection and its organization, we want to have a more \emph{effective} and \emph{efficient} retrieval solution, where we can achieve \textit{retrieve-everything-all-at-once}.
Inspired by~\cite{chen2024table}, which can be considered as a retrieve-all-tables-at-once solution that leverages a solver to reason about table relationships, we integrate its reasoning module as a way to \textit{search} for potential alignment to the data collection for an LLM. 

Intuitively, we instruct an LLM to reason about required data objects, augment it with a search process powered by a reasoning solver to figure out potential alignments to the data collection, and verify them through self-evaluation.

In the rest of the paper, we provide an overview of our methodology in Section~\ref{sec:method}, and elaborate each component of our solution in Section~\ref{sec:retrieve}. We evaluated our solution on two datasets, i.e., Bird and OTT-QA in Section~\ref{sec:exp}, and conclude in Section~\ref{sec:con}.  
\section{Overview}
~\label{sec:method}
This section provides an overview of our alignment-oriented LLM-based retrieval solution. Each component will then be elaborated in Section~\ref{sec:retrieve}. 
Intuitively, our solution adopts the idea of retrieving while generating ~\cite{riches}, so that we can take the most advantage of LLM's reasoning capability, and the decoding process enables an optimization of each component jointly through beam search. 
Rather than simply interleaving ``retrieval'' by constrained decoding based on evidence from the text corpus, and ``thoughts'' by unconstrained decoding~\cite{riches}, we propose to retrieve everything jointly through one decoding process guided by information from data objects available in the data collection, a solver's reasoning, and LLM's self-verification.

Specifically, we consider the retrieval problem as a generative problem that an LLM needs to output a ``reasoning process'' to find all necessary data objects from the data collection to answer the given question.
There are two main challenges for an off-the-shelf LLM to output such a reasoning process that can align with the data organization from the collection.
First, while an LLM can use its reasoning capabilities to analyze what information might be useful, it cannot determine how to map that information to existing data objects without access to the data collection.
Secondly, LLMs may lack the domain-specific knowledge to reason about
how different data objects that are semantically related to the question are connected. They might also struggle to identify whether additional data objects are required to connect these objects, particularly when the question involves private databases~\cite{chen2024beaver}. Therefore, LLMs need guidance from the data collection and its organization to generate the complete ``reasoning process''.

To solve the challenges mentioned above, we formulate the reasoning process generation as a decoding process consisting of multiple intermediate steps. Rather than traditional text decoding where each step decodes a token, we consider each step as an alignment that decodes a sequence of tokens based on the information of existing objects from the data collection.
We mainly consider two alignment steps with one self-verification step. The first is \emph{information alignment}, where we draft the key information needed to answer the question directly. This is achieved by constrained decoding using N-grams (introduced in Section ~\ref{sec:index}) from existing data objects.
The second is \emph{structure alignment}, where we reason about how different pieces of key information from existing data objects can be connected, potentially with additional objects, to answer the question through a reasoning solver.
The alignment results are then fed to the ``reasoning process'' as drafts, and the LLM self-verifies the relatedness of the data objects to the question as well as their connections and selects the final data objects that can fully answer the question.
The overall idea is illustrated in Figure~\ref{fig:problem} (rightmost block in upper half), omitting the details of beam search.
\section{Methodology}
\label{sec:retrieve}

\subsection{Indexing}
\label{sec:index}
In this paper, we unify both tables and passages (and potentially objects of other modalities such as images) and consider them as \textit{textual data objects}. We chunk each serialized data object, compute the embeddings of each chunk, and further represent and index it with an N-gram collection.

\paragraph{Serialization.}
In our data collection, each passage chunk is serialized as a concatenation of its passage title and the content of the chunk. A table chunk is serialized as a concatenation of its table name, title, description (if any), and rows.

\paragraph{N-Grams and Embeddings.}
N-grams are used to summarize the key information of a chunk from data objects. They can provide a quick lookup of what is contained in a chunk. Additionally, embeddings are used to support semantic similarity search. Both embeddings and N-grams are used to guide LLMs' reasoning of what data objects should be used to answer the question (Section~\ref{sec:decomp}).
Specifically, we constructed N-grams for each table and passage chunk, varying N between one to three. Moreover, we computed the embedding for each table and passage chunk.

\subsection{Alignment}
\subsubsection{Information Alignment}
\label{sec:decomp}
As described in Section~\ref{sec:method}, we instruct the LLM to generate a reasoning process with multiple intermediate steps. The first step is to determine the key information required to answer the question.
As an off-the-shelf LLM does not have access to the data collection, its analysis of what information might be helpful may not align with the information from the available data objects. To address this, we propose to instruct the LLM to first decompose the question by extracting keywords independently of the data collection, and then guide it to rephrase each keyword using N-grams available from the data objects in the collection through constrained decoding.

\paragraph{Constrained Beam Decoding.}
Our constrained-decoding-based information alignment is based on the model's extracted keywords from the user question. 
Since these keywords might not appear as directly in the corpus, we instruct the model to rephrase them to align with N-grams indexed from our data collection as mentioned in Section \ref{sec:index}.

This alignment is performed by constraining the model's output space during the decoding process.
The constrained beam decoding starts once a `(' is decoded which indicates that model is performing alignment and continues until `)' is decoded which indicates that the model has finished alignment for one keyword.
As an N-gram can be composed of multiple tokens, we use a suffix tree to keep track of valid continuations of generated tokens. With beam search, we maintain top lists of N-grams with the highest scores and decode the list with the highest score. The score of an N-gram is calculated as the average logits across all of its tokens.

Each list of N-grams decoded is used as a query to search for chunks using BM25. 
Additionally, for each object, we compute the embedding similarity between the user question and the serialized object (mentioned in Section \ref{sec:index}). The final score for each object is calculated as a weighted sum between the BM25 score and the embedding similarity. This score is then used to identify the most relevant objects, forming a \textit{base} set of search objects that serve as the foundation for constructing a draft for the continuation of LLM's ``reasoning process''.

\subsubsection{Structure Alignment}
\label{sec:ilp}

Information alignment guides us towards a set of search objects from the data collection that is very likely to help answer the question. However, there can be \emph{redundant} and \emph{missing} information. 
For instance, several passages identified may address the same aspect of the question. The information about the necessary bridging entities~\cite{yang2018hotpotqa} or bridging tables~\cite{chen2024table}, and the information that has to be derived from those bridging entities or tables
can still be missing.
To address this, we further design a structure alignment module that reasons about a complete list of search objects with their organization, so that it can match the information required and fully answer the question.

The key challenge of structure alignment is LLM does not have a global view of all available data objects and may lack specific knowledge to identify the missing objects needed to correctly connect the candidate objects that have been identified, especially when the data collection belongs to a specialized domain that the LLM may not have encountered extensively during its training.

Therefore, we propose to use an external solver to solve this structure alignment problem, where we can formulate the objective and include any business or domain-specific reasoning logic. Specifically, for a given list of search objects, the solver is tasked with returning a subset of the input search objects that are \textit{connected} and can be combined to answer the question. We then use the content of these selected objects to construct a partial ``draft'' for the LLM to continue its ``reasoning process''(Section~\ref{sec:verify}).


\paragraph{Inference using Mixed-Integer Programming.}
We formulate structure alignment for retrieval as an optimization problem. As outlined above, the goal is to identify a list of $k$ objects from a given list of $M$ search objects $\{O_{i}\}_{i=1}^{M}$ that can be connected to best answer the question $Q$ considering the domain-specific knowledge. Similar to~\cite{chen2024table}, we formulate it as a mixed-integer linear program (MIP) problem due to its flexibility to inject any business or domain-specific logic into the objective.

Specifically, the goal of the MIP program is to select a list of $k$ objects that can simultaneously maximize the relevance between the question and selected objects and compatibility (strength of connection) among the selected objects.

\textit{Relevance.} Relevance between a user question $Q$ and an object $O_i$, denoted as $R_i$, is defined as the cosine similarity between the embedding of $Q$ and the embedding of the serialized object $O_i$.

\textit{Compatibility.}
Object-object compatibility between two objects $O_i$ and $O_j$, denoted as $C_{ij}$, measures the strength of connection between two objects.
It is computed based on both semantic similarity (cosine similarity of embeddings) and exact-value similarity between contents from both objects. Details can be found in Appendix \ref{app:compatibility}.

Finally, we combine the above relevance and compatibility scores in the following MIP formulation with decision variables, objectives, and constraints. The output of the MIP consists of both the $k$ selected objects as well as the connections between these objects (connecting entities or joinable columns).

\noindent \textit{Decision variables.} To incorporate the relevance score, we define the binary decision variable $b_i$ to denote whether object $O_i$ is selected. To take into account the compatibility scores, we define the binary decision variable $c_{ij}$ to denote whether object $O_i$ is selected to connect with object $O_j$.

\noindent \textit{Objective.} Using the above variables and parameters, the objective function is to maximize the relevance of the selected objects and compatibility among the selected objects.
\begin{equation}
\arg \max \sum_i{R_i}b_i + \sum_{i,j}C_{ij}c_{ij}
\end{equation}

The objective function is subject to the following constraints:

\noindent \textit{Constraint 1} ensures decision variables are binary: $b_i, c_{ij} \in \{0, 1\}$



\noindent \textit{Constraint 2} sets the maximum number of objects and encourages selected objects to be connected: $\sum_{i} b_i = k, \sum_{i,j}c_{ij} \leq 2(k - 1) $


\noindent \textit{Constraint 3} ensures that only connections between selected objects are considered: $2c_{ij} \leq o_i + o_j, \forall i, j$


\paragraph{Constructing Multiple Alignment Drafts.}
Though our reasoning solver is to reason about the alignment that considers data organization to fully answer the question, it depends on the base set of data objects detected in the information alignment stage. Yet, this set of objects may not include sufficient information to answer the user question. In particular, for questions that require multiple connected objects (and bridging objects), the base objects might need to be expanded to include sufficient information. To handle this, we propose to expand the base search objects in different ways to \textit{multiple} sets of search objects and run the aforementioned MIP solver for each set of search objects to obtain \textit{multiple} drafts that combine both information and structure alignment. In our experiments, we considered three ways of expansions. 

We expand the base set of objects iteratively, controlling the number of steps and objects added per step. For each base object and each step, we can add its $k$ most compatible objects (calculated using the same compatibility score defined above). We can repeat the above process for $l$ steps, leading to more objects introduced.

\subsection{Self-Verfication and Aggregation}
\label{sec:verify}
Finally, from each draft generated by the MIP solver, the LLM itself will select a set of objects that can potentially be used to answer the user question, acting as a verifier that collectively evaluates all information decoded so far to perform object selection. An aggregation mechanism, based on model confidence, is used to combine the selection results from the different drafts.


\paragraph{Draft Injection.}
To inject a draft into the decoding process of an LLM, it is first serialized as a string, then models are enforced to decode this string using constrained decoding. As mentioned in Section \ref{sec:ilp}, a draft from MIP consists of both selected objects and connections among selected objects.
Each object is serialized in the same way described in Section \ref{sec:index}, but only the $k$ ($k=5$ in our setting) most similar table rows or passage sentences are selected, so that the model can focus on the most relevant content. 
Each connection is serialized as ``column \{column name\} in \{table name\} connects with column \{column name\} in \{table name\}'' to represent a joinable column or ``\{cell\} in \{table name\} connects with \{sentence\} in \{passage name\}'' to represent a connecting entity.

\paragraph{Model as Self-Verifier.}
At this stage, model has generated a decomposition of the user question, and has alignment knowledge on both information and structure. Using these knowledge, model performs the verification process by checking if selected objects includes content that can cover different aspects of the decomposition and that selected objects are connected. We use constraint decoding to guide this selection process to ensure factuality. In particular, we ensure that objects chosen by models must be present in the draft.

\paragraph{Aggregation Through Beam-Search}
After verification is performed for each draft, the model's \textit{reasoning process} is completed.
Since the \textit{reasoning process} is produced by a three-step constrained decoding process (different from the typical text decoding process where each step consists of a single token), and both information and structure alignment steps yield multiple outputs (sequences of tokens), we use beam search to generate multiple \textit{reasoning processes}.
We aggregate the selected data objects from the multiple beams by factoring in the model's confidence for each selected object.
Specifically, the aggregation process is treated as a weighted voting process, where each beam acts as a voter who votes for a selected object. In this context, an object's confidence can be measured by the weight of votes and the number of votes it receives.
The weight of a vote is measured using logits. In particular, tokens generated in the \textit{reasoning process} that correspond to the object's name are identified, and the logits of these tokens are averaged to be the weight of a vote.
The number of votes an object receives is the count of its occurrences across all beams, with softmax applied to normalize this count.
Then, the confidence in an object is computed as the weighted sum of the average vote weight and the normalized number of votes.
Finally, the data objects with the highest confidence are selected as the final set of retrieved objects.


\section{Experiments}
\label{sec:exp}

\subsection{Experimental setup}
\label{exp:setup}

\paragraph{Datasets.}
We evaluate our approach and baselines on open-domain question answering tasks that involve multiple sources of information. Therefore, we use OTT-QA~\cite{chen2020open} and Bird~\cite{li2024can}.  Specifically, OTT-QA involves questions on both passages and tables whose answers are mostly short text, while Bird involves questions on (multiple) tables, and the answer is a SQL statement.
For each dataset, we use the dev split for the user questions and constructing the data collection.
For Bird, we construct the data collection by merging tables from all databases used by the dev questions.
Similarly, for OTT-QA, we construct the data collection by merging the tables and passages used to answer the questions in the dev set.
In our experiments, we remove the questions we found with incomplete annotations (missing either required tables or passages through our manual inspection) from OTT-QA.
In total, there are 1834 questions and a collection of 3862 objects (3073 passages and 789 tables) for OTT-QA and 1534 questions and 75 tables in the data collection for Bird. After chunking (described in Section \ref{sec:index}), there are 4407 chunks and 249515 chunks for OTT-QA and Bird, respectively.

\paragraph{Baselines.}
We evaluate our approach against two baseline methods: the standard RAG and agentic RAG approaches. For the standard RAG baseline, we consider two variations: dense retrieval and dense retrieval followed by a cross-encoder reranker. Additionally, we enhance the standard RAG approach by incorporating an LLM-based query decomposition. To have a fair comparison with our approach, we use the same model (Llama 3.1-8B-Instruct) as the LLM to conduct query decomposition. Table \ref{tab:prompt-decomp} contains prompts used for generating sub-questions.

In our experiments, the embedding model used for dense retrieval was UAE-Large-V1 \cite{li2023angle}, and we use bge-reranker-v2-minicpm-layerwise \cite{li2023making, chen2024bge} as the reranking model. Additionally, ReAct was chosen as the representative of the agentic RAG approach.

All objects are chunked and serialized in the way described in Section \ref{sec:index}. The dense retrieval method computed the embedding for each chunk and outputs the top-$k$ objects with the highest cosine similarity with the user question. If an object is divided into multiple chunks, its similarity score is the highest similarity score across all its chunks.
For the reranking model, it was provided with the top-50 objects retrieved using the dense retrieval method. The reranker model assigns a score for each pair of user question and object, and it outputs top-$k$ objects with the highest scores. When query decomposition is applied, 30 objects were retrieved for each sub-question using the dense retrieval method and further reranked to output the top-$k$ objects.

ReAct was implemented following the original design of interleaving thought, action, and observation. A \textit{thought} step allows models to reason about the current situation. An \textit{action} can be of two types: (1) the model can generate some keywords to search for relevant objects from the corpus (2) or finish generation with an answer. An \textit{observation} step involves calling a dense retriever, which retrieves the 5 serialized objects with the highest similarity to the model-generated keywords in \textit{action}. Because most questions in both datasets can be answered using 4 objects, we set the maximum number of iterations to 8. The process continues until either the answer is found or the maximum limit of 8 rounds is reached.
Table \ref{tab:prompt-react} contains the 3-shot prompts used for ReAct.

\paragraph{Environment.}
We use the Python-MIP\footnote{\url{https://www.python-mip.com/}} package with Gurobi as the external Mixed-integer program solver and a cluster of V100 GPUs for performing inference.
Llama-3.1-8B-Instruct was used as the model for running \sys{} to generate the retrieval process. To perform the downstream tasks, we use Llama-3.1-8B-Instruct and GPT-4o-mini.
ReAct was executed on the same two models. The same three ICL examples were provided to models for running \sys{} and downstream tasks.
Table \ref{tab:prompt-arm} contains the 3-shot prompts used for \sys{}.

\subsection{Metrics}
We evaluate both the retrieval performance of the retrieved objects and the end-to-end performance on downstream tasks. We further report the number of LLM calls used for retrieval as a signal to reflect the cost.

For retrieval performance, we adopt the standard metrics of precision, recall, and F1 of the retrieved objects compared to the gold objects. However, we note that the recall metric could be misleading because in an extreme scenario, a retriever can achieve high recall by retrieving a significant portion of gold objects for every question, but with none of the questions having \textit{all} gold objects retrieved. This is problematic as a question can usually only be answered when all information provided. Therefore, we further augment existing metrics with the the percentage of questions with all gold objects retrieved, denoted as perfect recall (PR). For ReAct, we examine its retrieval performance by comparing the objects provided to the LLM with the gold objects.

For the end-to-end performance on downstream tasks, for OTT-QA, we compare the predicted short answer and gold short answer using exact match and F1 score. For Bird, following \cite{li2024can}, we compare the predicted SQL and gold SQL statement as the execution accuracy (1 if the execution results of both SQL statements are the same and 0 otherwise).

Regarding the number of LLM calls for retrieval,
\sys{} makes one LLM call as the entire retrieval process is completed in one decoding process. In ReAct, each LLM call produces an \textit{action} that formulates queries for retrieval. However, the retrieval results from the last call are not fed back into the LLM. As a result, the number of LLM calls for retrieval in ReAct is calculated as the total iterations minus one, which we assume the last call is for generating the final response.

\begin{table*}
\centering

\caption{Retrieval performance of baselines. PR is the percentage of questions with all gold objects retrieved. X-D refers to method X with query decomposition.}
\begin{adjustbox}{max width=\linewidth}
\begin{tabular}{ccccc|cccc|cccc|cccc}
& \multicolumn{4}{c}{Dense retrieval (DR)} & \multicolumn{4}{c}{DR + reranker (DRR)} & \multicolumn{4}{c}{DR-D} & \multicolumn{4}{c}{DRR-D}\\
\cmidrule(lr){2-5} \cmidrule(lr){6-9} \cmidrule(lr){10-13} \cmidrule(lr){14-17}
& P & R & F1 & PR & P & R & F1 & PR & P & R & F1 & PR & P & R & F1 & PR\\
\midrule

Bird  \\

\midrule

@2 & 53.5 & 58.7 & 54.1 & 34.3 & 60.5 & 65.8 & 61.1 & 43.5 & 52.4 & 48.5 & 48.0 & 25.1 & 59.8 & 65.4 & 60.5 & 44.2\\
@3 & 45.1 & 73.2 & 54.1 & 52.4 & 47.5 & 76.0 & 56.8 & 58.4 & 46.3 & 59.4 & 49.4 & 35.7 & 47.4 & 76.2 & 56.7 & 59.0\\
@5 & 33.7 & 89.0 & 47.7 & 78.4 & 33.0 & 86.2 & 46.5 & 74.7 & 38.9 & 74.9 & 49.2 & 56.5 & 33.0 & 86.3 & 46.6 & 74.6\\
\midrule

OTT-QA \\
\midrule

@2 & 62.2 & 52.4 & 55.1 & 20.6 & 71.5 & 60.7 & 63.7 & 32.3 & 61.2 & 47.4 & 51.3 & 13.4 & 70.6 & 60.0 & 63.0 & 32.2\\
@3 & 48.3 & 59.7 & 51.4 & 30.0 & 54.7 & 68.2 & 58.5 & 43.5 & 46.7 & 54.2 & 48.1 & 22.2 & 54.1 & 67.5 & 57.9 & 42.7\\
@5 & 34.6 & 69.0 & 44.0 & 43.1 & 37.4 & 75.2 & 47.8 & 53.8 & 32.7 & 62.3 & 40.9 & 32.7 & 37.1 & 74.5 & 47.4 & 52.5\\

\bottomrule
\end{tabular}
\end{adjustbox}

\label{tab:retrieval-baseline}
\end{table*}

\begin{table}
\centering
\small
\caption{Retrieval performance of \sys{} and ReAct. For ReAct, \textit{Avg \#obj.} refers to the average number of retrieved objects throughout all LLM calls. For \sys{}, \textit{Avg \#obj.} refers to the average number of it outputs. \#calls refers to the number of LLM calls used for retrieval. \sys{} was performed on Llama3.1-8B-Instruct.}
\begin{adjustbox}{max width=\linewidth}
\begin{tabular}{ccccccc}
& \#calls $\downarrow$ & Avg \#obj. $\downarrow$ & P & R & F1 & PR\\
\midrule
Bird  \\

\midrule

\makecell{ReAct\\(Llama3.1-8B-It)} & 5.26 & 17.3 & 15.0 & 96.7 & 24.5 & 93.5 \\
\noalign{\vskip 0.5ex}
\makecell{ReAct\\(GPT4o-mini)} & 3.16 & 11.4 & 25.1 & 97.0 & 37.8 & 93.3\\
\noalign{\vskip 0.5ex}
\sys{}&1 & 5.00 & 42.7 & 96.5 & 56.0 & 92.7\\

\midrule
OTT-QA \\
\midrule

\makecell{ReAct\\(Llama3.1-8B-It)} & 4.52 & 19.5 & 15.3 & 76.0 & 23.1 & 55.1 \\
\noalign{\vskip 0.5ex}
\makecell{ReAct\\(GPT4o-mini)} & 3.68 & 14.5 & 21.7 & 80.6 & 30.9 & 62.7 \\
\noalign{\vskip 0.5ex}
\sys{}& 1& 4.98 & 47.3 & 79.8 & 55.0 & 62.5\\

\bottomrule
\end{tabular}
\end{adjustbox}
\label{tab:retrieval-r2}
\end{table}

\subsection{Retrieval performance}
Table \ref{tab:retrieval-baseline} shows the retrieval performance of dense retrievers and dense retrievers with reranker on Bird and OTT-QA. Table \ref{tab:retrieval-r2} shows the retrieval performance of ReAct and \sys{}. As mentioned in Section \ref{exp:setup}, the retrieval process of \sys{} was performed by Llama3.1-8B-Instruct in one LLM call.

On Bird, \sys{} retrieves on average 5.00 objects, achieving a recall of 96.5 and perfect recall of 92.7. In comparison, the best-performing standard RAG baseline, dense retrieval, retrieves 5 objects with a recall of 89.0 and perfect recall of 78.4, which is 7.5 and 14.3 points lower compared to \sys{}, respectively. Additionally, compared to ReAct running on Llama3.1-8B-Instruct, \sys{} reduces LLM calls by 4.26 and retrieves 12.3 fewer objects while maintaining comparable recall and perfect recall but achieving a 31.5 higher F1, potentially reducing noise.

On OTT-QA, \sys{} retrieves an average of 4.98 objects, achieving a recall of 79.8 and a perfect recall of 62.5. In comparison, the best-performing standard RAG baseline, dense retrieval with reranker, retrieves 5 objects with a recall of 75.2 and a perfect recall of 53.8, which is 4.6 and 8.7 points lower than \sys{}, respectively. Additionally, compared to ReAct running on Llama3.1-8B-Instruct, \sys{} reduces LLM calls by 3.52 and retrieves 14.5 fewer objects while achieving 3.8 points higher recall and 7.4 points higher perfect recall.

These results demonstrate that \sys{} outperforms standard RAG baselines in recall and perfect recall, indicating a higher likelihood of retrieving all necessary information to answer user questions. Additionally, compared to agentic RAG, \sys{} achieves comparable or superior retrieval performance while using fewer LLM calls and retrieving fewer objects.

\subsection{End-to-end performance}

Table \ref{tab:ete} shows the end-to-end results on both datasets across all approaches and two models.

Averaging across both models, \sys{} demonstrates superior performance on Bird and OTT-QA. On Bird, it outperforms the best-performing standard RAG baseline, dense retrieval, by 2.55 points in execution accuracy and agentic RAG by 11.1 points. On OTT-QA, \sys{} achieves 3.7 points higher exact match and 4.4 points higher F1 match compared to dense retrieval with reranker, while outperforming the agentic RAG by 12.7 points in exact match and 14.6 points in F1 match.

The results indicate that \sys{} outperforms standard RAG baselines by retrieving objects of higher quality, leading to improved downstream performance despite retrieving a similar number of objects. Moreover, compared to agentic RAG baselines, \sys{} also achieves superior downstream performance using fewer LLM calls. This highlights \sys{} as a more effective and efficient solution.

\subsection{ReAct Analysis}

We randomly selected 50 questions from each of the Bird and OTT-QA datasets and manually analyzed the results generated by both models using the ReAct approach. Our analysis showed two primary types of errors made by models during its iterative reasoning process. First, models might forget information it generated in previous iterations.
Secondly, models can fall into cycles of searching for similar keywords, even when relevant objects have already been retrieved. Both behaviors can lead to inefficiency and potentially a large number of LLM calls. Detailed examples can be found in Appendix \ref{app:react-bad}.

\begin{table}
\centering
\caption{End-to-end execution accuracy of all methods.}

\begin{adjustbox}{max width=\linewidth}
\begin{tabular}{cccc|ccc}
& \multicolumn{6}{c}{Generation model}\\
\cmidrule(lr){2-7}
& \multicolumn{3}{c}{Llama3.1-8B-It} & \multicolumn{3}{c}{GPT4o-mini}\\
\cmidrule(lr){2-4} \cmidrule(lr){5-7}
& \multicolumn{1}{c}{Bird} & \multicolumn{2}{c}{OTT-QA} & \multicolumn{1}{c}{Bird} & \multicolumn{2}{c}{OTT-QA}\\
\cmidrule(lr){2-2} \cmidrule(lr){3-4} \cmidrule(lr){5-5} \cmidrule(lr){6-7}
& Acc  & Exact & F1 & Acc & Exact & F1\\
\midrule
DR@5 & 17.5 & 34.2 & 40.6 &  29.7 & 40.0 & 47.8 \\
DRR@5 & 16.8 & 39.8 & 46.9 & 29.8 & 45.9 & 55.2 \\
DR-D@5 & 13.9 & 27.2 & 33.3 & 23.8 & 32.6 & 39.7 \\
DRR-D@5 & 15.9 & 38.8 & 46.3 & 26.5 & 45.6 & 55.4 \\
ReAct & 4.7 & 27.0 & 32.5 & 25.4 & 40.7 & 49.3 \\
\sys{} & \textbf{20.6} & \textbf{44.0} & \textbf{51.8} & \textbf{31.7} & \textbf{49.1} & \textbf{59.1}\\

\bottomrule
\end{tabular}
\end{adjustbox}

\label{tab:ete}
\end{table}







\subsection{Ablation studies}
\begin{figure}
\centering
\includegraphics[width=\linewidth]{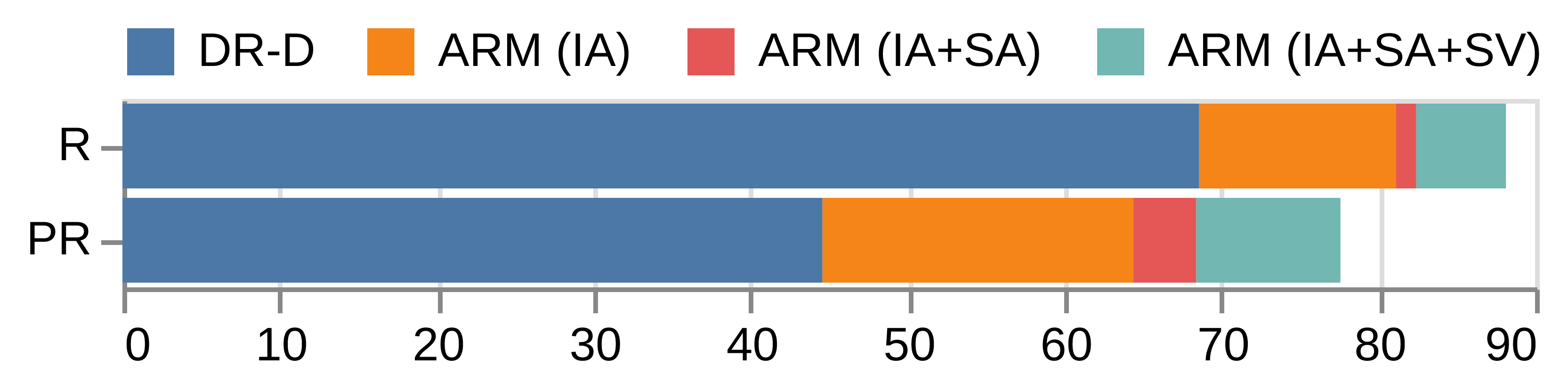}
\caption{The average recall and perfect recall of dense retrieval with decomposition (DR-D) and our method with successive modules: information alignment (IA), structure alignment (SA), and self-verification and aggregation (SV) for the top-5 retrieved objects across Bird and OTT-QA.}
\label{fig:ablation}
\end{figure}

\paragraph{Significance of different modules.}
\sys{} includes three modules: information alignment, structure alignment, and self-verification and aggregation. Figure \ref{fig:ablation} illustrates the performance of dense retrieval with query decomposition and our method with successive modules for the top-5 retrieved objects. Information alignment involves decomposing the original question into keywords and retrieving relevant objects, similar to the baseline of dense retrieval with query decomposition. To highlight the benefits of information alignment, we compare the two methods. On average, information alignment outperforms dense retrieval with query decomposition by 12.5 points in recall and 19.8 points in perfect recall across two datasets.
The inclusion of structure alignment boosts recall by 1.28 points and perfect recall by 4.02 points, building on the gains from information alignment. Finally, the complete method with all three modules enhances recall by 5.72 points and perfect recall by 9.18 points. The improvement with each successive module demonstrates the contribution of every module to the overall retrieval performance.

\begin{figure}
\centering
\includegraphics[width=\linewidth]{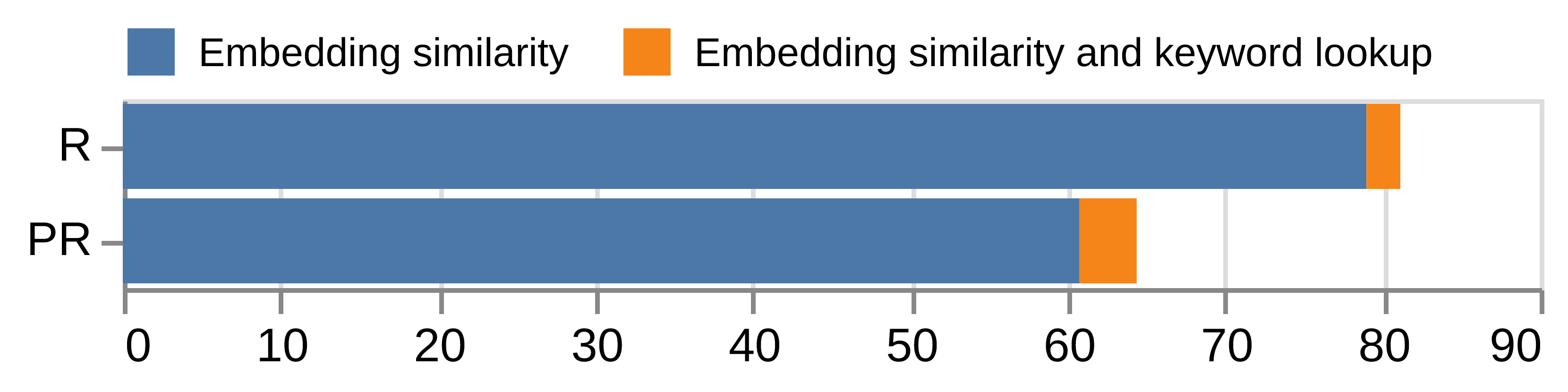}
\caption{The average recall and perfect recall for the information alignment module evaluated using embedding similarity alone and when combined with keyword lookup for the top-5 retrieved objects across Bird and OTT-QA.}
\label{fig:ablation_ia}
\end{figure}

\paragraph{Information alignment.} Information alignment retrives relevant objects through two components: keyword lookup using the decomposed and aligned keywords via BM25, and embedding similarity. Figure \ref{fig:ablation_ia} illustrates the performance with embedding similarity alone and when combined with keyword lookup. Adding keyword lookup to embedding similarity improves recall by 2.15 points and perfect recall by 3.65 points on average across two datasets, clearly demonstrating the contribution of each component.
\section{Conclusion}
~\label{sec:con}
Understanding what data objects are available in the data collection and their organization is critical for answering complex open-domain questions that involve heterogeneous information sources. Query decomposition by an off-the-shelf LLM generates queries without an awareness of what is available in the data collection, often leading to a suboptimal retrieval performance. Although agentic RAG can interact with the data collection, the queries are formulated based on previous retrieval results rather than an understanding of the available data objects and their organization. Therefore, agentic RAG is often inefficient to retrieve all required data objects due to more LLM calls.
In this work, we propose an alignment-oriented retrieval method \sys{}, that is capable of exploring which data objects may contain the key information required to answer the question, and also navigating the data organization to identify all required data objects, even when their information is not explicitly stated in the question.
Our experimental evaluations showed that, compared to the baselines, \sys{} is more effective in terms of performance in retrieval and downstream tasks, as well as more efficient in terms of the number of LLM calls needed.

\bibliography{custom}

\begin{thebibliography}{21}
\providecommand{\natexlab}[1]{#1}

\bibitem[{Asai et~al.(2023)Asai, Wu, Wang, Sil, and Hajishirzi}]{self-rag-2023}
Akari Asai, Zeqiu Wu, Yizhong Wang, Avirup Sil, and Hannaneh Hajishirzi. 2023.
\newblock Self-rag: Learning to retrieve, generate, and critique through self-reflection.
\newblock \emph{arXiv preprint arXiv:2310.11511}.

\bibitem[{Chen et~al.(2024{\natexlab{a}})Chen, Xiao, Zhang, Luo, Lian, and Liu}]{chen2024bge}
Jianlv Chen, Shitao Xiao, Peitian Zhang, Kun Luo, Defu Lian, and Zheng Liu. 2024{\natexlab{a}}.
\newblock \href {https://arxiv.org/abs/2402.03216} {Bge m3-embedding: Multi-lingual, multi-functionality, multi-granularity text embeddings through self-knowledge distillation}.
\newblock \emph{Preprint}, arXiv:2402.03216.

\bibitem[{Chen et~al.(2024{\natexlab{b}})Chen, Wenz, Zhang, Kayali, Tatbul, Cafarella, Demiralp, and Stonebraker}]{chen2024beaver}
Peter~Baile Chen, Fabian Wenz, Yi~Zhang, Moe Kayali, Nesime Tatbul, Michael Cafarella, {\c{C}}a{\u{g}}atay Demiralp, and Michael Stonebraker. 2024{\natexlab{b}}.
\newblock Beaver: An enterprise benchmark for text-to-sql.
\newblock \emph{arXiv preprint arXiv:2409.02038}.

\bibitem[{Chen et~al.(2024{\natexlab{c}})Chen, Zhang, and Roth}]{chen2024table}
Peter~Baile Chen, Yi~Zhang, and Dan Roth. 2024{\natexlab{c}}.
\newblock Is table retrieval a solved problem? exploring join-aware multi-table retrieval.
\newblock In \emph{Proceedings of the 62nd Annual Meeting of the Association for Computational Linguistics (Volume 1: Long Papers)}, pages 2687--2699.

\bibitem[{Chen et~al.(2020)Chen, Chang, Schlinger, Wang, and Cohen}]{chen2020open}
Wenhu Chen, Ming-Wei Chang, Eva Schlinger, William Wang, and William~W Cohen. 2020.
\newblock Open question answering over tables and text.
\newblock \emph{arXiv preprint arXiv:2010.10439}.

\bibitem[{Jain et~al.(2024)Jain, Soares, and Kwiatkowski}]{riches}
Palak Jain, Livio~Baldini Soares, and Tom Kwiatkowski. 2024.
\newblock From rag to riches: Retrieval interlaced with sequence generation.
\newblock \emph{arXiv preprint arXiv:2407.00361}.

\bibitem[{Jhamtani et~al.(2023)Jhamtani, Fang, Xia, Levy, Andreas, and Van~Durme}]{jhamtani2023natural}
Harsh Jhamtani, Hao Fang, Patrick Xia, Eran Levy, Jacob Andreas, and Ben Van~Durme. 2023.
\newblock Natural language decomposition and interpretation of complex utterances.
\newblock \emph{arXiv preprint arXiv:2305.08677}.

\bibitem[{Khot et~al.(2022)Khot, Trivedi, Finlayson, Fu, Richardson, Clark, and Sabharwal}]{khot2022decomposed}
Tushar Khot, Harsh Trivedi, Matthew Finlayson, Yao Fu, Kyle Richardson, Peter Clark, and Ashish Sabharwal. 2022.
\newblock Decomposed prompting: A modular approach for solving complex tasks.
\newblock \emph{arXiv preprint arXiv:2210.02406}.

\bibitem[{Kojima et~al.(2022)Kojima, Gu, Reid, Matsuo, and Iwasawa}]{kojima2022large}
Takeshi Kojima, Shixiang~Shane Gu, Machel Reid, Yutaka Matsuo, and Yusuke Iwasawa. 2022.
\newblock Large language models are zero-shot reasoners.
\newblock \emph{Advances in neural information processing systems}, 35:22199--22213.

\bibitem[{Li et~al.(2023)Li, Liu, Xiao, and Shao}]{li2023making}
Chaofan Li, Zheng Liu, Shitao Xiao, and Yingxia Shao. 2023.
\newblock \href {https://arxiv.org/abs/2312.15503} {Making large language models a better foundation for dense retrieval}.
\newblock \emph{Preprint}, arXiv:2312.15503.

\bibitem[{Li et~al.(2024)Li, Hui, Qu, Yang, Li, Li, Wang, Qin, Geng, Huo et~al.}]{li2024can}
Jinyang Li, Binyuan Hui, Ge~Qu, Jiaxi Yang, Binhua Li, Bowen Li, Bailin Wang, Bowen Qin, Ruiying Geng, Nan Huo, et~al. 2024.
\newblock Can llm already serve as a database interface? a big bench for large-scale database grounded text-to-sqls.
\newblock \emph{Advances in Neural Information Processing Systems}, 36.

\bibitem[{Li and Li(2023)}]{li2023angle}
Xianming Li and Jing Li. 2023.
\newblock Angle-optimized text embeddings.
\newblock \emph{arXiv preprint arXiv:2309.12871}.

\bibitem[{Press et~al.(2022)Press, Zhang, Min, Schmidt, Smith, and Lewis}]{self-ask-2023}
Ofir Press, Muru Zhang, Sewon Min, Ludwig Schmidt, Noah~A Smith, and Mike Lewis. 2022.
\newblock Measuring and narrowing the compositionality gap in language models.
\newblock \emph{arXiv preprint arXiv:2210.03350}.

\bibitem[{Shinn et~al.(2024)Shinn, Cassano, Gopinath, Narasimhan, and Yao}]{shinn2024reflexion}
Noah Shinn, Federico Cassano, Ashwin Gopinath, Karthik Narasimhan, and Shunyu Yao. 2024.
\newblock Reflexion: Language agents with verbal reinforcement learning.
\newblock \emph{Advances in Neural Information Processing Systems}, 36.

\bibitem[{Trivedi et~al.(2022)Trivedi, Balasubramanian, Khot, and Sabharwal}]{ircot}
Harsh Trivedi, Niranjan Balasubramanian, Tushar Khot, and Ashish Sabharwal. 2022.
\newblock Interleaving retrieval with chain-of-thought reasoning for knowledge-intensive multi-step questions.
\newblock \emph{arXiv preprint arXiv:2212.10509}.

\bibitem[{Wei et~al.(2022)Wei, Wang, Schuurmans, Bosma, Xia, Chi, Le, Zhou et~al.}]{wei2022chain}
Jason Wei, Xuezhi Wang, Dale Schuurmans, Maarten Bosma, Fei Xia, Ed~Chi, Quoc~V Le, Denny Zhou, et~al. 2022.
\newblock Chain-of-thought prompting elicits reasoning in large language models.
\newblock \emph{Advances in neural information processing systems}, 35:24824--24837.

\bibitem[{Yang et~al.(2018)Yang, Qi, Zhang, Bengio, Cohen, Salakhutdinov, and Manning}]{yang2018hotpotqa}
Zhilin Yang, Peng Qi, Saizheng Zhang, Yoshua Bengio, William~W Cohen, Ruslan Salakhutdinov, and Christopher~D Manning. 2018.
\newblock Hotpotqa: A dataset for diverse, explainable multi-hop question answering.
\newblock \emph{arXiv preprint arXiv:1809.09600}.

\bibitem[{Yao et~al.(2024)Yao, Yu, Zhao, Shafran, Griffiths, Cao, and Narasimhan}]{yao2024tree}
Shunyu Yao, Dian Yu, Jeffrey Zhao, Izhak Shafran, Tom Griffiths, Yuan Cao, and Karthik Narasimhan. 2024.
\newblock Tree of thoughts: Deliberate problem solving with large language models.
\newblock \emph{Advances in Neural Information Processing Systems}, 36.

\bibitem[{Yao et~al.(2022)Yao, Zhao, Yu, Du, Shafran, Narasimhan, and Cao}]{react}
Shunyu Yao, Jeffrey Zhao, Dian Yu, Nan Du, Izhak Shafran, Karthik Narasimhan, and Yuan Cao. 2022.
\newblock React: Synergizing reasoning and acting in language models.
\newblock \emph{arXiv preprint arXiv:2210.03629}.

\bibitem[{Zhang et~al.(2024)Zhang, Liao, Li, and Du}]{zhang2024agentic}
Weinan Zhang, Junwei Liao, Ning Li, and Kounianhua Du. 2024.
\newblock Agentic information retrieval.
\newblock \emph{arXiv preprint arXiv:2410.09713}.

\bibitem[{Zhou et~al.(2022)Zhou, Sch{\"a}rli, Hou, Wei, Scales, Wang, Schuurmans, Cui, Bousquet, Le et~al.}]{zhou2022least}
Denny Zhou, Nathanael Sch{\"a}rli, Le~Hou, Jason Wei, Nathan Scales, Xuezhi Wang, Dale Schuurmans, Claire Cui, Olivier Bousquet, Quoc Le, et~al. 2022.
\newblock Least-to-most prompting enables complex reasoning in large language models.
\newblock \emph{arXiv preprint arXiv:2205.10625}.

\end{thebibliography}

\appendix

\section{Compatibility}
\label{app:compatibility}
\textit{Table-table compatibility} is determined by pairwise column comparisons between the two tables. Since only one pair of columns is necessary to connect two tables, table compatibility is determined by the highest compatibility score among all possible column pairs. Each column-column compatibility is calculated as the weighted sum of the semantic similarity between the column headers and the exact-value similarity (Jaccard similarity) of the column rows.

\textit{Table-passage compatibility} is determined by comparing all cells within a table and all sentences in a passage and taking the pair with the highest compatibility. Each cell-sentence compatibility is calculated as the weighted sum of the semantic similarity and exact-value similarity (overlap coefficient) between the cell content and the sentence.

\textit{Passage-passage compatibility} is computed by comparing all sentences in a passage with all sentences in the other passage and taking the pair with the the highest compatibility. Each sentence-sentence compatibility is calculated as the weighted sum of semantic similarity and exact-value similarity (overlap coefficient).

\section{Prompts}
Prompts for our experiments are included in Table \ref{tab:prompt-decomp} - \ref{tab:prompt-arm}.

\begin{table*}
\centering

\caption{3-shot prompt used for query decomposition.}

\begin{adjustbox}{max width=\linewidth}
\begin{tabular}{p{\linewidth}}

\toprule
You are given a user question, your task is to decompose the user question into simpler sub-questions. The sub-questions should be separated by newline characters.\\
Here are some examples.\\
\\
User question: what is the full name of the jesus college alumni who graduated in 1960 ?\\
Sub-questions:\\
Who are the alumni of Jesus College?\\
Which alumni of Jesus College graduated in 1960?\\
What are the full names of those who graduated from Jesus College in 1960?\\
\\
User question: the home stadium of the bray wanderers of 2006 league of ireland is situated behind what station ?\\
Sub-questions:\\
What is the home stadium of the Bray Wanderers in the 2006 League of Ireland?\\
Where is the Bray Wanderers' home stadium located geographically?\\
Which station is situated in front of this stadium?\\
What is the name of the station that is located behind the home stadium of the Bray Wanderers?\\
\\
User question: what is the enrollment of the school with a delta sigma theta chapter chartered on may 8 , 1971 ?\\
Sub-questions:\\
What is Delta Sigma Theta?\\
Which schools have a Delta Sigma Theta chapter?\\
Which schools have a Delta Sigma Theta chapter chartered on May 8, 1971?\\
What is the enrollment of the school that has a Delta Sigma Theta chapter chartered on May 8, 1971?\\
\\
User question: \{user question\}\\
\bottomrule
\end{tabular}
\end{adjustbox}

\label{tab:prompt-decomp}
\end{table*}

\begin{table*}
\centering

\caption{3-shot prompt used for ReAct.}

\begin{adjustbox}{max width=\linewidth}
\small
\begin{tabular}{p{\linewidth}}

\toprule

Solve a question answering task with interleaving Thought, Action, Observation steps. A Thought step must be followed by a Action step, a Action step must be followed by a Observation step, and a Observation step must be followed by a Thought step. Thought can reason about the current situation, and Action can be of two types: \\
(1) Search[keyword], which returns five new documents or tables relevant to the keywords.\\
(2) Finish[answer], which returns the answer and finishes the task.\\
You answer should be a short answer to the user question.\\
Here are some examples.\\
\\
User question: what is the full name of the jesus college alumni who graduated in 1960 ?\\
Thought: I need to search jesus college alumni, find the people that graduated in 1960, then find their full names\\
Action: Search[jesus college aumni]\\
Observation: \\
(...5 tables omitted...)\\
\\
Thought: The person who graduated in 1960 is not mentioned in any table. So I need to see more content relevant to jesus college alumni to decide.\\
Action: Search[jesus college alumni]\\
Observation: \\
(...5 tables omitted...)\\
\\
Thought: The person who graduated in 1960 is not mentioned in any table. So I need to see more content relevant to jesus college alumni to decide.\\
Action: Search[jesus college alumni]\\
Observation: \\
(...2 tables omitted...)\\
Table id: List\_of\_politicians,\_lawyers,\_and\_civil\_servants\_educated\_at\_Jesus\_College,\_Oxford\_1\\
Table name: List of politicians, lawyers, and civil servants educated at Jesus College, Oxford\\
Table description: Jesus College is one of the constituent colleges of the University of Oxford in England. ...\\
Table content:\\
| Name                  |    M |    G | Degree                                               | Notes                                                                                                                                           |\\
|:----------------------|-----:|-----:|:-----------------------------------------------------|:------------------------------------------------------------------------------------------------------------------------------------------------|\\
| Lalith Athulathmudali | 1955 | 1960 | BA Jurisprudence ( 2nd , 1958 ) , BCL ( 2nd , 1960 ) | President of the Oxford Union ( 1958 ) ; a Sri Lankan politician ; killed by the Tamil Tigers in 1993                                           |\\
| Neal Blewett ( HF )   | 1957 | 1959 | BA PPE ( 2nd )                                       | Member of the Australian House of Representatives ( 1977-1994 ) , Government Minister ( 1983-1994 ) , High Commissioner to the UK ( 1994-1998 ) |\\
(...2 tables omitted...)\\
\\
Thought: Table List\_of\_politicians,\_lawyers,\_and\_civil\_servants\_educated\_at\_Jesus\_College,\_Oxford\_1 lists Lalith Athulathmudali as receiving his BCL degree in 1960 from jesus college. So the person who graduated in 1960 is Lalith Athulathmudali. I need to search Lalith Athulathmudali to find their full name.\\
Action: Search[Lalith Athulathmudali]\\
Observation: \\
Document id: /wiki/Lalith\_Athulathmudali\\
Document name: /wiki/Lalith\_Athulathmudali\\
Document content: Lalith William Samarasekera Athulathmudali , PC ( Sinhala :  ; 26 November 1936 - 23 April 1993 ) , known as Lalith Athulathmudali , was Sri Lankan statesman . He was a prominent member of the United National Party , who served as Minister of Trade and Shipping ; Minister National Security and Deputy Minister of Defence ; Minister of Agriculture , Food and Cooperatives and finally Minister of Education . Following a failed impeachment of President Premadasa , he was removed from the UNP and formed his own party .\\
(...4 documents omitted...)\\
\\
Thought: Document Lalith\_Athulathmudali provides the full name: Lalith William Samarasekera Athulathmudali. So the answer is Lalith William Samarasekera Athulathmudali.\\
Action: Finish[Lalith William Samarasekera Athulathmudali]\\
\\
(...2 examples omitted...)\\
\\
User question: \{user question\}\\
\bottomrule
\end{tabular}
\end{adjustbox}

\label{tab:prompt-react}
\end{table*}

\begin{table*}
\centering

\caption{3-shot prompt used for \sys{}.}

\begin{adjustbox}{max width=\linewidth}
\small
\begin{tabular}{p{\linewidth}}

\toprule
You are given a user question, your task is to decompose the user question into contiguous, non-overlapping substrings that can cover different information mentioned in the user question. For each substring, generate n-grams that are the most relevant to the substring.
Based on the generated relevant n-grams, generate a list of relevant objects, including their names, content, and connections between these objects. From these candidate objects, you should identify the minimum number of objects that can be used to answer the user question based on the relevance between the object name, object content and user question as well as the relevance of the object connections. You should end your response with <>.\\
\\
User question: what is the full name of the jesus college alumni who graduated in 1960 ?\\
The relevant keywords are full name | jesus college | alumni | graduated | 1960\\
The relevant n-grams are full name ( name) | jesus college ( jesus college) | alumni ( alumni, former, university) | graduated ( degree, educated, postgraduate) | 1960 ( 1960)\\
\\
Here are the objects that can be relevant:\\
(...4 tables omitted...)\\
Table id: List\_of\_politicians,\_lawyers,\_and\_civil\_servants\_educated\_at\_Jesus\_College,\_Oxford\_1\\
Table name: List of politicians, lawyers, and civil servants educated at Jesus College, Oxford\\
Table description: Jesus College is one of the constituent colleges of the University of Oxford in England. ...\\
Table content:\\
| Name                  |    M |    G | Degree                                               | Notes                                                                                                                                           |\\
|:----------------------|-----:|-----:|:-----------------------------------------------------|:------------------------------------------------------------------------------------------------------------------------------------------------|\\
| Lalith Athulathmudali | 1955 | 1960 | BA Jurisprudence ( 2nd , 1958 ) , BCL ( 2nd , 1960 ) | President of the Oxford Union ( 1958 ) ; a Sri Lankan politician ; killed by the Tamil Tigers in 1993                                           |\\
| Neal Blewett ( HF )   | 1957 | 1959 | BA PPE ( 2nd )                                       | Member of the Australian House of Representatives ( 1977-1994 ) , Government Minister ( 1983-1994 ) , High Commissioner to the UK ( 1994-1998 ) |\\
(...2 documents omitted...)\\
Document id: /wiki/Lalith\_Athulathmudali\\
Document name: /wiki/Lalith\_Athulathmudali\\
Document content: Lalith William Samarasekera Athulathmudali , PC ( Sinhala :  ; 26 November 1936 - 23 April 1993 ) , known as Lalith Athulathmudali , was Sri Lankan statesman . He was a prominent member of the United National Party , who served as Minister of Trade and Shipping ; Minister National Security and Deputy Minister of Defence ; Minister of Agriculture , Food and Cooperatives and finally Minister of Education . Following a failed impeachment of President Premadasa , he was removed from the UNP and formed his own party .\\
(...2 documents omitted...)\\
\\
Here are the potential connections between these objects:\\
List\_of\_politicians,\_lawyers,\_and\_civil\_servants\_educated\_at\_Jesus\_College,\_Oxford\_1 includes Lalith Athulathmudali which connects with /wiki/Lalith\_Athulathmudali\\
(...4 connections omitted...)\\
\\
Here I summarize the relevant information to answer the user question\\
<table> List\_of\_politicians,\_lawyers,\_and\_civil\_servants\_educated\_at\_Jesus\_College,\_Oxford\_1 </table> lists Lalith Athulathmudali as receiving his BCL degree in 1960 from jesus college, which covers the keywords "jesus college alumni graduated 1960" | <document> /wiki/Lalith\_Athulathmudali </document> extends on this by providing his full name: Lalith William Samarasekera Athulathmudali, which covers the keywords "full name" <>\\
\\
(...2 examples omitted...)\\
\\
User question: \{user question\}\\
\bottomrule
\end{tabular}
\end{adjustbox}

\label{tab:prompt-arm}
\end{table*}

%







%







\section{Examples where models fail using ReAct}
\label{app:react-bad}

Below are examples where models failed to generate correct answers using ReAct.

As seen in Table \ref{tab:react-forget}, the model can forget information generated in previous iterations. It was trying to search the population of Barcelos, but concluded with the population of Ajim.

\begin{table*}
\centering

\caption{Example where models forget information generated in previous iterations using ReAct.}

\begin{adjustbox}{max width=\linewidth}
\small
\begin{tabular}{p{\linewidth}}

\toprule

User question: how many people live in the city where the first ipsc european handgun championship , junior category , event took place ?\\
\\
Thought: I need to search the first IPSIC European Handgun Championship, Junior Category, to find the event location, then search the city to find its population.\\
Action: Search[IPSC European Handgun Championship, Junior Category]\\
Observation:\\
Table id: IPSC\_European\_Handgun\_Championship\_2\\
Table name: IPSC European Handgun Championship\\
Table description: The IPSC European Handgun Championship is an IPSC level 4 championship hosted every third year in Europe.\\
Champions -- Junior category\\
\\
Table content:\\
|   Year | Division   | Gold             | Silver                     | Bronze           | Venue                  |\\
|-------:|:-----------|:-----------------|:---------------------------|:-----------------|:-----------------------|\\
|   2013 | Open       | Simo Partanen    | Daniil Karchev             | Francois Belloni | Barcelos , Portugal    |\\
|   2013 | Standard   | Kenneth Handberg | Sotirios-Thomas Zafeiridis | Mats Selven      | Barcelos , Portugal    |\\
|   2013 | Production | Pavel Torgashov  | Konstantin Kryuchin        | Nikita Kryuchin  | Barcelos , Portugal    |\\
|   2016 | Standard   | Danila Pakhomov  | Ilya Sologub               | Kirill Fedorov   | Felsőtárkány , Hungary |\\
(...4 objects omitted...)\\
\\
Thought: Table IPSC\_European\_Handgun\_Championship\_2 lists the venue for 2013 event in Barcelos, Portugal. So the event took place in Barcelos, Portugal. I need to search Barcelos to find its population.\\
Action: Search[\textbf{Barcelos}]\\
(...retrieved objects omitted...)\\
\\
Thought: The content of the documents includes no information about population. I need to search Barcelos to find its population.\\
Action: Search[\textbf{Barcelos}]\\
(...retrieved objects omitted...)\\
\\
Thought: The content of the documents includes no information about population. I need to search Barcelos to find its population.\\
Action: Search[\textbf{Barcelos}]\\
(...retrieved objects omitted...)\\
\\
Thought: The content of the documents includes no information about population. I need to search Barcelos to find its population.\\
Action: Search[\textbf{Barcelos}]\\
Observation:\\
(...four objects omitted...)\\
Document id: /wiki/Ajim\\
Document name: /wiki/Ajim\\
Document content: Ajim ( Arabic : Aǧīm ) is a commune and port located on the Island of Djerba off the coast of Tunisia . It is Djerba 's main fishing port and the closest city to the African continent . It had a population of 24,294 at the 2014 census .\\
\\
Thought: Document \textbf{/wiki/Ajim} lists a population of 24,294 at the 2014 census. So the population is 24,294.\\
Action: Finish[24,294]\\
\bottomrule
\end{tabular}
\end{adjustbox}

\label{tab:react-forget}
\end{table*}

As seen in Table \ref{tab:react-loop}, the model fell into a loop of searching using similar keywords, even when gold tables have already been retrieved.
The gold tables are \texttt{financial.card}, \texttt{financial.disp}, \texttt{financial.client}

\begin{table*}
\centering

\caption{Example where models loop on similar keywords using ReAct. Gold tables are bolded.}

\begin{adjustbox}{max width=\linewidth}
\small
\begin{tabular}{p{\linewidth}}

\toprule
User question: provide the ids and age of the client with high level credit card, which is eligible for loans.\\
\\
Thought: I need to find tables that can cover client ids, age, credit card level, and loan eligibility. So I should search for client information, credit card details, and loan eligibility status.\\
Action: Search[client information, credit card details, loan eligibility]\\
Observation:\\
Table name: \textbf{financial.client}\\
|   client\_id | gender   | birth\_date   |   district\_id |\\
|------------:|:---------|:-------------|--------------:|\\
|        1039 | F        | 1980-10-11   |            70 |\\
\\
Table name: financial.loan\\
|   loan\_id |   account\_id | date       |   amount |   duration |   payments | status   |\\
|----------:|-------------:|:-----------|---------:|-----------:|-----------:|:---------|\\
|      7264 |        11141 | 1997-10-07 |    44940 |         12 |       3745 | A        |\\
\\
Table name: \textbf{financial.disp}\\
|   disp\_id |   client\_id |   account\_id | type      |\\
|----------:|------------:|-------------:|:----------|\\
|       525 |         525 |          436 | DISPONENT |\\
\\
Table name: debit\_card\_specializing.customers\\
|   CustomerID | Segment   | Currency   |\\
|-------------:|:----------|:-----------|\\
|         4643 | SME       | EUR        |\\
\\
Table name: debit\_card\_specializing.products\\
|   ProductID | Description                 |\\
|------------:|:----------------------------|\\
|         544 | Guarantee Credit Note       |\\
\\
Thought: The tables I have found do not include information about credit card levels or loan eligibility. I need to search for more specific tables that might contain credit card details and loan eligibility status.\\
Action: Search[credit card level, loan eligibility status]\\
Observation:\\
Table name: financial.loan\\
|   loan\_id |   account\_id | date       |   amount |   duration |   payments | status   |\\
|----------:|-------------:|:-----------|---------:|-----------:|-----------:|:---------|\\
|      7308 |        11362 | 1996-12-27 |   129408 |         24 |       5392 | A        |\\
\\
Table name: \textbf{financial.client}\\
|   client\_id | gender   | birth\_date   |   district\_id |\\
|------------:|:---------|:-------------|--------------:|\\
|        1039 | F        | 1980-10-11   |            70 |\\
\\
Table name: debit\_card\_specializing.products\\
|   ProductID | Description           |\\
|------------:|:----------------------|\\
|         544 | Guarantee Credit Note |\\
\\
Table name: \textbf{financial.card}\\
|   card\_id |   disp\_id | type   | issued     |\\
|----------:|----------:|:-------|:-----------|\\
|       162 |       967 | junior | 1996-12-23 |\\
\\
Table name: debit\_card\_specializing.customers\\
|   CustomerID | Segment   | Currency   |\\
|-------------:|:----------|:-----------|\\
|          123 | SME       | EUR        |\\
\\
Thought: The tables I have found still do not include information about credit card levels or loan eligibility. I need to search for more specific tables that might contain credit card details and loan eligibility status.\\
Action: Search[credit card level, loan eligibility status]\\
(...retrieved tables omitted...)\\
\\
Thought: I still haven't found the necessary information regarding credit card levels and loan eligibility. I need to search for more specific tables that might contain this information.\\
Action: Search[credit card level, loan eligibility status] \\
...\\
\bottomrule
\end{tabular}
\end{adjustbox}

\label{tab:react-loop}
\end{table*}

\end{document}